\documentclass{article}

\usepackage{microtype}
\usepackage{graphicx}
\usepackage{subfigure}
\usepackage{booktabs} % for professional tables

\usepackage{amsfonts}       % blackboard math symbols
\usepackage{nicefrac}       % compact symbols for 1/2, etc.
\usepackage{microtype}      % microtypography
\usepackage{todonotes}
\usepackage{amsmath}
\usepackage{bm}
\usepackage{tikz}
\usepackage{pgfplots}
\usepackage{xcolor}
\usetikzlibrary{decorations.pathmorphing, shapes.misc, calc}
\definecolor{C1}{HTML}{1F77B4}
\definecolor{C2}{HTML}{FF7F0E}
\definecolor{mLightBrown}{HTML}{EB811B}
\definecolor{mLightGreen}{HTML}{14B03D}

\usepackage{hyperref}

\usepackage[accepted]{icml2020}

\icmltitlerunning{Rodent: Relevance determination in differential equations}

\newcommand{\x}{\bm{x}}

\newcommand{\z}{\bm{z}}
\newcommand{\mz}{\bm{\mu}_{z}}

\newcommand{\se}{\sigma_e}
\newcommand{\sz}{\bm{\sigma_z}}
\newcommand{\laz}{\bm{\lambda}_z}

\newcommand{\X}{\bm{X}}

\newcommand{\N}{\mathcal{N}}
\newcommand{\U}{\mathcal{U}}
\newcommand{\E}[2]{\text{E}_{#1}\left[#2\right]}

\begin{document}

\twocolumn[
\icmltitle{Rodent: Relevance determination in differential equations}



\icmlsetsymbol{equal}{*}

\begin{icmlauthorlist}
  \icmlauthor{Niklas Heim}{cvut}
  \icmlauthor{V\'aclav \v Sm\'idl}{cvut}
  \icmlauthor{Tom\'a\v s Pevn\'y}{cvut}
\end{icmlauthorlist}

\icmlaffiliation{cvut}{Czech Technical University, Prague, Czechia}
\icmlcorrespondingauthor{Niklas Heim}{niklas.heim@aic.fel.cvut.cz}

\icmlkeywords{Bayesian, ARD, ODE, differential equation, relevance determination, deep learning}

\vskip 0.3in
]



\printAffiliationsAndNotice{\icmlEqualContribution} % otherwise use the standard text.

\begin{abstract}
  We aim to identify the generating, ordinary differential equation (ODE) from
  a set of trajectories of a partially observed system.  Our approach does not
  need prescribed basis functions to learn the ODE model, but only a rich set of
  \emph{Neural Arithmetic Units}.
  For maximal explainability of the
  learnt model, we minimise the state size of the ODE as well as the
  number of non-zero parameters that are needed to solve the problem.
  This sparsification is realized through a combination of the
  \emph{Variational Auto-Encoder} (VAE) and \emph{Automatic Relevance
  Determination} (ARD). We show that it is possible to learn not only one
  specific model for a single process, but a manifold of models representing
  harmonic signals as well as a manifold of Lotka-Volterra systems.
\end{abstract}

\section{Introduction}%
\label{sec:introduction}

Many real-world dynamical systems lack a mathematical description because it is
too complicated to derive one from first principles such as symmetry or
conservation laws. To tackle this problem, we introduce a novel combination of
a sparsity-promoting  \emph{Variational Auto-Encoder} (VAE) and an
\emph{ordinary differential equation} (ODE) solver that can discover governing
equations in the form of nonlinear ODEs. Moreover, our goal is not just to
build a model for a specific ODE, but to discover the simplest mathematical
formula for a manifold of ODEs parametrized by few parameters, e.g. harmonic
oscillators with varying frequency $\omega$.  Estimating parameters from data
requires learning a mapping from the input space to a (hidden) parameter space,
which makes the choice of an autoencoder natural.

Finding mathematical expressions from data is often referred to as equation
discovery, a prototypical approach is called SINDy~\cite{kaiser_sparse_2018}.
SINDy learns a sparse selection of terms from an extensive library of basis
functions, which makes it easily interpretable. But unlike us, most other
approaches focus on learning a model for a single process and not for a
manifold.
 
The rest of this section provides an informal introduction to the problem of
our interest.  A more formal problem definition and the description of our
relevance determination in ODEs (Rodent) is given in
Sec.~\ref{sec:problem_definition} \& \ref{sec:rodent}.
\begin{figure}
  \centering
  \includegraphics[width=\linewidth]{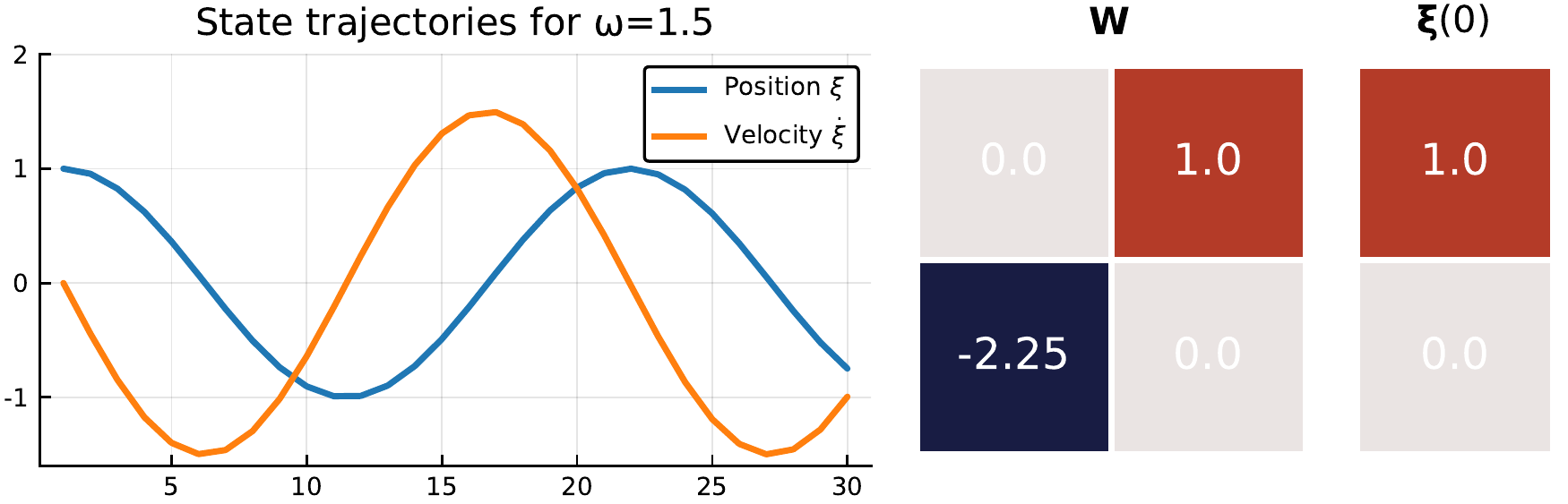}
  \caption{State trajectories of a harmonic oscillator with frequency
  $\omega=1.5$ on the left (position in blue, velocity in yellow). The RHS of
  the generating ODE (as defined in Eq.~\eqref{eq:harmonic_system}) on the
  right.}
  \label{fig:perfect_harmonic_nau}
\end{figure}

An educating example for learning right-hand sides of ODEs is learning the
manifold of harmonic oscillators. They are generated by solving the
second-order ODE
\begin{equation}
  \label{eq:harmonic_2nd_order}
  \ddot\xi = -\omega^2 \xi,
\end{equation}
with position $\xi,$ acceleration $\ddot\xi$ (second time derivative), and
frequency $\omega$ of the oscillator.  Since any ODE of order $N$ can be
written as a system of $N$ first order ODEs, Eq.~\eqref{eq:harmonic_2nd_order}
becomes
\renewcommand*{\arraystretch}{1.3}
\begin{align}
  \label{eq:harmonic_system}
  \frac{\partial\bm\xi}{\partial t} =
  \begin{bmatrix}
   \dot{\xi} \\ \ddot{\xi}
  \end{bmatrix}
  &=
  \begin{bmatrix}
    0 & 1 \\
    -\omega^2 & 0
  \end{bmatrix}
  \begin{bmatrix}
    \xi \\ \dot{\xi}
  \end{bmatrix}
  = \bm W \bm\xi,
\end{align}
with the ODE model $\bm W$ and the state $\bm\xi.$ The ODE state is composed of
the oscillator's position and its velocity.
Fig.~\ref{fig:perfect_harmonic_nau} shows the state trajectories of a harmonic
oscillator with frequency $\omega=1.5$ and the ODE (RHS of
Eq.~\eqref{eq:harmonic_system}) that generates them.  In practice, it is rare
to observe the complete state of a dynamical system, which motivates us to
learn the governing ODE from partial observations.  Thus, our goal becomes to
learn the RHS of Eq.~\eqref{eq:harmonic_system} only by observing a set of
trajectories $\{\x_i\}_{i=1}^l$ of the position $\xi$ (blue curve in
Figure~\ref{fig:perfect_harmonic_nau})
\begin{equation}
  \label{eq:x_trajectory}
  \x_i = [\xi_1,\xi_2,\ldots,\xi_K],
\end{equation}
where each $\x_i$ is assumed to be generated from an oscillator with a
different frequency.  The downside of having access only to partial
observations of $\bm\xi$ we cannot enforce the second state variable  to be the
velocity of the oscillator, i.e. the solution of the system is not unique.  For
example any transformation $\bm T = [1 \; 0; a \; b]$, where $a,b$ are free
parameters, of Eq.~\eqref{eq:harmonic_system} with transformed state
$\bm\xi^{*}$ and transformed model $\bm W^{*}$
\begin{align}
  \label{eq:transform_W}
  \bm W^{*}  &= \bm T \bm W \bm T^{-1}
             &&= \begin{bmatrix} -a/b & 1/b \\ -\omega^2 b - a/b & a/b \end{bmatrix}\\
  \label{eq:transform_xi}
  \bm\xi^{*} &= \bm T \bm \xi
             &&= \begin{bmatrix} \xi \\ a\xi + b\dot\xi \end{bmatrix}
\end{align}
is a possible solution of Eq.~\eqref{eq:harmonic_system}. In order to find
solutions that are expected and understandable by humans, we apply
\emph{automatic relevance determination} (ARD; Sec.~\ref{sec:rodent}) to the
ODE model and the state parameters.

\subsection{Contributions}%
\label{sub:contributions}
\begin{itemize}
  \item Our relevant ODE identifier (\emph{Rodent}) consists of a VAE with ARD
    prior on the latent dimension and an ODE solver as the VAE decoder, which
    is, to the best of our knowledge, has not been used before.

  \item A successfully trained Rodent is very explainable and able to output a
    human readable ODE:
    \begin{equation}
      \text{explain}(\text{Rodent}, [\sim]\:\text{- data})
          \rightarrow \ddot \xi = -\omega^2\xi + \ldots
    \end{equation}
    We achieve this by building the ODE model with sparse \emph{Neural
    Arithmetic Units} (Sec.~\ref{sec:neural_arithmetic}). We do not need a
    large library of predefined basis functions (as required by e.g.
    \citealt{kaiser_sparse_2018}), or prior physical
    knowledge~\cite{raissi_physics_2017}. We achieve sparsity by replacing the
    L1 regularization of the neural arithmetic units by ARD, which was shown
    yield better results~\cite{wipf_new_2018}.

  \item Our encoder is split into two parts:
    i) A dense neural network (NN) receiving the first few steps of the
    time-series estimates the full ODE state $\bm\xi$ from partial
    observations $\x_i$.
    ii) A convolutional NN that is agnostic to the length of the time-series
    learns ODE parameters $\bm W$.

  \item The system can learn not only solutions to one specific process but a manifold
    of solutions, e.g. to harmonic signals with a frequency parametrizing the manifold.
\end{itemize}

\section{Problem definition}%
\label{sec:problem_definition}

We are concerned with models of  time-series
\begin{equation}
  \X_i = [\bm x_1, \bm x_2,\ldots, \bm x_K]
\end{equation}
with $\bm x_k \in \mathbb{R}^d$ that are generated by discrete-time, noisy
observations of a continuous-time process
\begin{equation}
  \bm x_k = H(\bm{\xi}(\Delta t k)) + e_k,
\end{equation}
where $k=1\ldots K$ and $e_k \sim \N(0,\se^2\mathbf{I})$. The partial
observation operator $H$ has fixed sampling intervals $\Delta t$, and the
temporal evolution of the state variable $\bm\xi(t) \in \mathbb{R}^n$ is
governed by a dynamical system described by an ODE:
\begin{align}
  \label{eq:dyn_sys}
  \frac{\partial \bm{\xi}}{\partial t} = & f(\bm{\theta}, t).
\end{align}
The solution of the ODE for given parameters $\bm\theta$ and initial conditions
$\bm\xi(0)$ is
\begin{equation}
  \bm{\xi}(t)=\psi(\bm\theta,\bm{\xi}(0),t),
\end{equation}
where $\psi$ is a standard ODE solver, such as Tsit5~\cite{tsitouras_runge-kutta_2011}.
We aim to learn the structure of the ODE model from a training set of $L$
trajectories $\{\X_i\}_{i=1}^{L}$  generated by the same generative process but
with  different parameters and different initial conditions, for each
trajectory, i.e.
\begin{align}
  \label{eq:series_x}
  \X_i =& H(\psi(\bm\theta_i,\bm{\xi}_i(0), \bm{t}))+\bm{e},
\end{align}
where $\bm{t} = [0, \Delta t, \ldots, K\Delta t]$.  Assuming we observe a
system with expected order $M$ and unknown structure, we choose $N \geq M$. Eq.
\eqref{eq:series_x} thus defines a generative model of sequences $\X_i$ from the
latent space of parameters $\bm{\theta}$ and initial conditions $\bm{\xi}(0)$.
Following the variational autoencoder approach \cite{kingma_auto-encoding_2013}, we define
the latent space $\bm{z}=[\bm\theta,\bm\xi(0)]$ with the ODE \eqref{eq:dyn_sys}
playing the role of the decoder. We seek an encoder in the form of a
distribution $q(\bm{z}|\x)$, parametrized by a deep neural network. Moreover,
we use a constrictive prior $p(\z)$ to promote sparsity on $\bm z$ to obtain a simple
model, which is important for the explainability of the learnt ODEs. 
We will use the normal-gamma prior proposed in \cite{neal_bayesian_1996} and
recently used e.g.  in Bayesian compression of Neural Networks
\cite{louizos_bayesian_2017, louizos_learning_2017}.

Our method is applicable to any differentiable dynamical system with trainable
parameters $\bm\theta$.  We will demonstrate our framework by learning
parameters to \emph{Neural Arithmetic} layers
(Sec.~\ref{sec:neural_arithmetic}) which aim to maximise the explainability of
the learnt manifolds.

\section{Rodent -- Relevant ODE identifier}%
\label{sec:rodent}

\begin{figure}
  \centering
  \resizebox{.49\textwidth}{!}{\def\baselen{2.6cm}

\begin{tikzpicture}[shorten >=1pt,->,draw=black!50, node distance=\baselen]
  \tikzstyle{every pin edge}=[<-,shorten <=1pt]
  \tikzstyle{neuron}=[circle,fill=black!25,minimum size=17pt,inner sep=0pt]
  \tikzstyle{input neuron}=[neuron, fill=gray];
  \tikzstyle{output neuron}=[neuron, fill=gray];
  \tikzstyle{hidden neuron}=[neuron, fill=mLightBrown];
  \tikzstyle{annot} = [text width=4em, text centered]

  % Draw the input layer nodes
  \foreach \name / \y in {1,...,5}
  % This is the same as writing \foreach \name / \y in {1/1,2/2,3/3,4/4}
    \node[input neuron] (I-\name) at (-3,-\y) {};

  \node[draw, ultra thick, rounded corners=8, minimum width=\baselen, minimum height=1.1cm]
    (dense) at ($(I-2) + (3,0)$) {
      \begin{tabular}{c}
        Dense part of $\phi_\omega$ \\ \emph{only first N steps}
      \end{tabular}
    };
  \node[draw, ultra thick, rounded corners=8, minimum width=\baselen, minimum height=1.5cm]
    (conv) at ($(I-4) + (3,0)$) {
      \begin{tabular}{c}
        Conv. part of $\phi_\omega$\\
        \emph{full time series}
      \end{tabular}
    };

  % Draw the hidden layer nodes
  \node[hidden neuron] (H-1) at (2.2cm,-2.5cm) {$\bm \xi$};
  \node[hidden neuron] (H-2) at (2.2cm,-3.5cm) {$\bm \theta$};

  % Connect input -> conv
  \foreach \source in {1,...,5}
    \path[thick] (I-\source) edge (conv);
  % Connect input -> dense
  \path[thick] (I-1) edge (dense);
  \path[thick] (I-2) edge (dense);

  % Connect networks -> latent
  \path[thick] (dense) edge (H-1);
  \path[thick] (conv) edge (H-2);

  % Draw the output layer nodes
  \foreach \name / \y in {1,...,5}
      \node[output neuron] (O-\name) at (\baselen*2.5,-\y cm) {};

  % draw and connect decoder
  \node[draw, ultra thick, rounded corners=8, minimum height=1.1cm, minimum width=2.5cm]
      (odesolve) at  (\baselen*1.6, -3) {
        \begin{tabular}{c}
          ODE solver\\
          $H(\psi(\bm \theta, \bm \xi))$
        \end{tabular}
      };
  \foreach \name / \y in {1,...,2}
      \path[thick] (H-\name) edge (odesolve);
  \foreach \dest in {1,...,5}
      \path[thick] (odesolve) edge (O-\dest);

  % input sines
  \draw[-, very thick, decoration={snake, segment length=0.32cm, amplitude=0.6cm}, decorate]
    ($(I-1) - (2, .5)$) -- ($(I-1) - (.5, .5)$);
  \draw[-, very thick, decoration={snake, segment length=1.3cm, amplitude=0.6cm}, decorate]
    ($(I-1) - (2, 2)$) -- ($(I-1) - (.5, 2)$);
  \draw[-, very thick, decoration={snake, segment length=0.7cm, amplitude=0.6cm}, decorate]
    ($(I-1) - (2, 3.5)$) -- ($(I-1) - (.5, 3.5)$);

  % output sines
  \draw[-,dotted, very thick, decoration={snake, segment length=0.32cm, amplitude=0.6cm}, decorate]
    ($(O-1) + (.5, -.5)$)--($(O-1) + (2, -.5)$);
  \draw[-,dotted, very thick, decoration={snake, segment length=1.3cm, amplitude=0.6cm}, decorate]
    ($(O-1) + (.5, -2)$)--($(O-1) + (2, -2)$);
  \draw[-,dotted, very thick, decoration={snake, segment length=0.7cm, amplitude=0.6cm}, decorate]
    ($(O-1) + (.5, -3.5)$)--($(O-1) + (2, -3.5)$);

  % Annotate with distributions
  \node[above of=I-3] (pzx) {$\mathcal{N}(\bm z | \phi_\omega(\bm x), \bm\sigma_z)$};
  \node[annot, below of=H-1] (pzx) {$\mathcal{N}(0, \bm \lambda_z)$};
  \node[above of=O-3] (pxz) {$\mathcal{N}(\bm x | H(\psi(\bm z)), \sigma_x)$};
\end{tikzpicture}}
  \caption{Schematic of the Rodent.  Inputs and outputs in grey, latent
  variable $\z=[\bm \theta, \bm\xi(0)]$ in the middle in brown.  The encoder
  network $\phi_\omega$ on the left represents the mean of the posterior.
  It consists of a dense part to estimate the initial conditions and a convolutional
  part that is responsible for the ODE parameters.
  By using convolutions for the parameter estimation, we enable the network to
  process time-series of variable lengths.
  The decoder on the right is a combination of an ODE solver and the
  observation operator $H$.
  The latent variable has the ARD prior $\mathcal{N}(0,\bm\lambda_z)$.
  }
  \label{fig:rodent_schematic}
\end{figure}
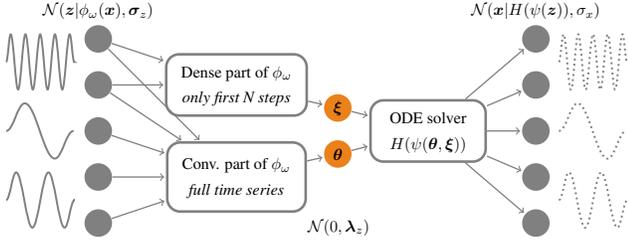

For simplicity of the notation we will write $\x=\text{vec}(\bm X_i)$ for the
flattened matrix of trajectories and we combine ODE state and parameters in a 
single latent variable $\z=[\bm\theta,\bm\xi(0)]$. Further we introduce a shorthand
for the ODE solver $\psi(\z) = H(\psi(\z,\bm t))$ (from Eq.~\eqref{eq:series_x}).
Now we can define the data likelihood as
\begin{align}
  \label{eq:decoder}
  p(\x|\z) &= \N(\x|\psi(\z),\sigma_x^2),
\end{align}
where the variance $\sigma_x^2$ is shared for all timesteps
$\bm t$ and datapoints $\x$.  By using an ODE solver (not a neural network) as
the decoder we enforce a structure in the latent space that allows for an
interpretation e.g. in terms of physical properties of the model.

To determine the structure of the ODE, we employ the \emph{Automatic Relevance
Determination} (ARD) prior (developed by \citealt{mackay_bayesian_1994,
neal_bayesian_1996}) on the latent layer:
\begin{align}
  \label{eq:ard_model}
  p(\z) &= \N(\z|0,\text{diag}(\laz^2)) &
  p(\laz) &= 1/\laz,%\Gamma(\laz|\alpha_0, \beta_0)
\end{align}
where a new vector variable $\laz>0$ of the same dimension as $\z$ has been
introduced. We will treat $\laz$ as an unknown variable and we will seek its
point estimate.

The approximated posterior distribution of the latent variable given the
observed sequence is prescribed by
\begin{align}
  \label{eq:ard_posterior}
  q(\z|\x) = \N(\z|\phi_\omega(\x), \bm{\sigma}_z^2),
\end{align}
where mean $\mz=\phi_\omega(\x)$ is a deep neural network with parameters
$\omega$, and both standard deviations $\laz$, and $\sz$ are shared for all
datapoints $\x$. Note that we apply ARD only to the ODE parameters and initial
conditions collected in $\z$ and not to the encoder network parameters
$\omega$.

The parameters of the posterior are obtained by
maximization of the \emph{Evidence Lower Bound} (ELBO) $\mathcal{L}$:
\begin{align}
  \mathcal{L} &= \E{p(\z|\x)}{\log p(\x|\z)} 
               - \text{KL}(q(\z|\x)||p(\z)). \label{eq:ELBO}
\end{align}
By applying the reparametrization trick and Monte-Carlo sampling inside the
expectation we can perform stochastic gradient descent on the ELBO:
\begin{equation}
\begin{aligned}
  \label{eq:elbo_reparam}
  \mathcal{L} &= \sum_{i=1}^n \E{}{\frac{(\x_i - \psi(\phi_\omega(\x_i) + \bm{\sigma}_z \odot \bm{\epsilon}))^2}{2\se^2}}
              + \frac{nd}{2}\log(\se) \\
              &+ \sum_{i=1}^n \left(
                  \log\left(\frac{\laz^2}{\sz^2}\right)
                  -m + \frac{\sz^2}{\laz^2} + \frac{\phi_\omega(\x_i)^2}{\laz^2}
              \right),
\end{aligned}
\end{equation}
with noise $\bm{\epsilon} \sim \N(0,I)$ and $\text{dim}(\z) = m$.  The ELBO is
maximized with respect to the parameters $\omega$ of the encoder network, and
the variances $\laz,\bm\sigma_z,\sigma_x$. The resulting algorithm is a
relevance determination of ODEs (\emph{Rodent}).

A schematic of the Rodent is shown in Fig.~\ref{fig:rodent_schematic}. The
encoder network consists of two parts. A dense network estimating the initial state
$\bm \xi(0)$ from the partial observations $\x$ receives only a few steps of
the beginning of the time-series since this part is the most relevant for
initial conditions. The second part of the encoder predicting the ODE
parameters $\bm \theta$ is a convolutional neural net (CNN). The CNN averages
over the time dimension after the convolutions, which allows it to use samples
of different length.

\section{Neural Arithmetic}%
\label{sec:neural_arithmetic}

Although it is theoretically possible to use an ODE model of arbitrary
complexity (conventional NN) to represent $f(\bm\theta,t)$, these models would
be difficult to understand for humans.  Therefore, we use stacks of sparse
neural arithmetic units (NMU and NAU are defined below) to represent nonlinear
ODEs with as few layers and parameters as possible.

\paragraph{Neural Multiplication Units}%
\label{par:neural_multiplication_units}
(NMU) as introduced by \citet{madsen_neural_2020} are capable of computing
products of inputs.  They use explicit multiplications and a gating mechanism,
which makes them easy to learn, but incapable of performing addition,
subtraction, and division.
We will use a Bayesian version of the NMU:
\begin{align}
  \label{eq:nmu}
  \hat{m}_{ij} &= \max(\min(m_{ij}, 1), 0) \\
  a_i    &= \prod_j \left( x_j \hat{m}_{ij} + 1 - \hat{m}_{ij} \right),
\end{align}
where the regularisation term proposed by \citet{madsen_neural_2020} is replaced 
by ARD acting on $m_{ij}$ (collected in a matrix $\bm
M$).

\paragraph{Neural Addition Units}%
\label{par:neural_arithmetic_units}
(NAU) can compute linear combinations (i.e. addition/subtraction). They can be
easily represented by a matrix multiplication (without a bias term)
\begin{equation}
  \bm a = \bm W \bm x,
\end{equation}
where the sparsification of $\bm W$ is again achieved through Bayesian
compression.

An NMU layer followed by an NAU layer can thus learn additions, subtractions,
and multiplications of inputs while still retaining maximal interpretability of
the resulting models. Given a stack of an NMU and an NAU with input
$\x=[x_1,x_2]$, NMU with $\bm M = [1 \; 0; 1 \: 1]$, and an NAU with $\bm W=[2,
1]$ we can even write a simple macro that outputs the resulting \LaTeX~equation:
\begin{equation}
  \text{NAU}(\text{NMU}(\x)) = 2x_1 + x_1x_2.
\end{equation}
Another interesting arithmetic unit is the \emph{Neural Arithmetic Logic Unit}
(NALU; \citealt{trask_neural_2018}). With a slight extension it can represent
addition, subtraction, multiplication, division, and arbitrary power functions.
Unfortunately, the NALU suffers from unstable convergence and cannot handle
negative inputs~\cite{madsen_neural_2020}, which makes it infeasible for our
purposes.

\section{Reidentification}%
\label{sec:reidentification}

A perfectly trained encoder will always output good parameters that reconstruct
the training sequence well. However, this task is complicated because
the encoder not only has to extract the initial conditions $\bm\xi(0)$ from
partial observations but also predict the ODE parameters $\bm\theta$ from a
manifold of generating processes.
It is likely that the encoder learns the manifold only approximately, which
means that estimated parameters $\z$ are not optimal for every datapoint $\x$.
Moreover, the generalization capability of the encoder outside the parameters
in the training data  will be limited.
To solve this, we introduce another optimization step after the Rodent's encoder
is trained, which we call \emph{reidentification}.

The goal of reidentification is to find the best latent code $\z$ for a given
input sample $\x$ by minimizing the error of the decoder\footnote{Note that the
encoder does not enter $\mathcal{R}$ and that the decoder does not have any
parameters.}. Specifically, we minimise the reconstruction error
\begin{align}
  \label{eq:reid_loss}
  \mathcal{R} = \text{MSE}(\psi(\z), \x)
\end{align}
with respect only to the relevant parameters of $\z$ (namely those with
non-zero mean or variance). 
The loss function~\eqref{eq:reid_loss} can be easily optimized using gradient 
based techniques (e.g. LBFGS or SGD), as the ODE solver is differentiable, and
we use samples of the latent code of the encoder $q(\z|\x)$ as starting 
points.

As will be shown in the experimental section (in
Fig.~\ref{fig:rodent_scores}), reidentification helps to 
extrapolate far beyond the range of parameters in the training data.

\section{Related work}%
\label{sec:related_work}

\paragraph{Discovering differential equations}%
\label{par:learning_odes}
Learning physically relevant concepts of the two-body problem was recently
solved both by \emph{Hamiltonian Neural Networks}
\cite{greydanus_hamiltonian_2019, toth_hamiltonian_2019,
sanchez-gonzalez_hamiltonian_2019}, which learn the Hamiltonian from
observations. Once the Hamiltonian is known, it is possible to derive from it
most other variables of interest.  While this is a very promising approach that
yields good results, it is limited to systems that can be described in terms of
positions and momenta such as interacting particles.\\
\citet{iten_discovering_2020} have shown how to discover the heliocentricity of
the solar system with an autoencoder, but their architectures make significant
prior assumptions, e.g. on the size of the latent space, which we do not need because
we employ the ARD prior.\\
Similarly, \citet{raissi_physics_2017} introduce physically informed NNs, which
perform well and are very interpretable if the general structure of the problem
is already known.\\
SINDy~\cite{kaiser_sparse_2018} is learning a few active terms from a library
of basis functions, which provides a very well interpretable result, but is
limited by the library of basis functions.\\
Another interesting approach is Koopman theory
\cite{rudy_data-driven_2017, brunton_discovering_2016} which relies on finding
a transformation into a higher dimensional space in which the ODE becomes linear.
Combining Koopman theory and autoencoders has produced promising results, such
as learning surrogates for a nonlinear pendulum and the Lorenz system
\cite{champion_data-driven_2019}. While the Koopman approach provides very
accurate predictions once its nonlinear transformation is found, it lacks
explainable results which we provide by learning the ODEs directly.

\paragraph{Bayesian compression}%
\label{par:bayesian_compression}
Recent work has shown that adopting a Bayesian
point of view to parameter pruning can significantly reduce computational cost
while still achieving competitive accuracy \cite{louizos_bayesian_2017,
louizos_learning_2017}.  These approaches focus on pruning network weights or
complete neurons to decrease computational cost and reduce
overfitting.  Our approach, however, enforces sparsity \emph{only on the latent
variable} and the main goal is interpretability of the latent variable.

\paragraph{Hyper networks} are neural nets that generate weights for another
network \cite{ha_hypernetworks_2016}. Our encoder can be seen as a kind of
hypernetwork, as it estimates the parameters of ODE, which can be seen as a
continuous-time residual neural network~\cite{chen2018neural}.

\section{Experiments}%
\label{sec:experiments}

In this section, we demonstrate how the Rodent learns two different systems.  We
begin by learning a simple manifold of harmonic signals
(Sec.~\ref{sub:identification_of_the_harmonic_oscillator}), two superimposed
harmonic signals, and then identify the equations of a nonlinear ODE called the
Lotka-Volterra (LK) system (Sec.~\ref{sub:lotka_volterra_system}).

\subsection{Identification of the harmonic oscillator}%
\label{sub:identification_of_the_harmonic_oscillator}
\begin{figure}
  \centering
  \includegraphics[height=.32\linewidth]{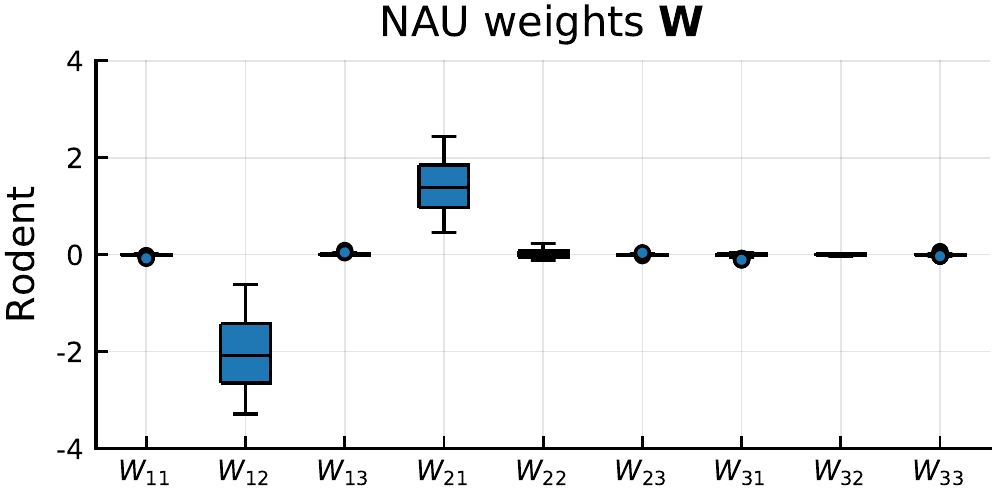}
  \includegraphics[height=.32\linewidth]{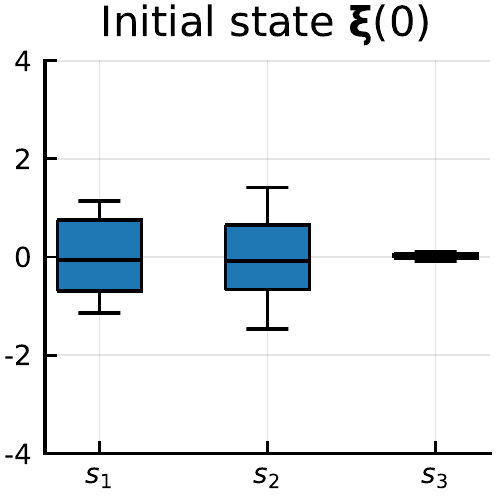}
  \includegraphics[height=.32\linewidth]{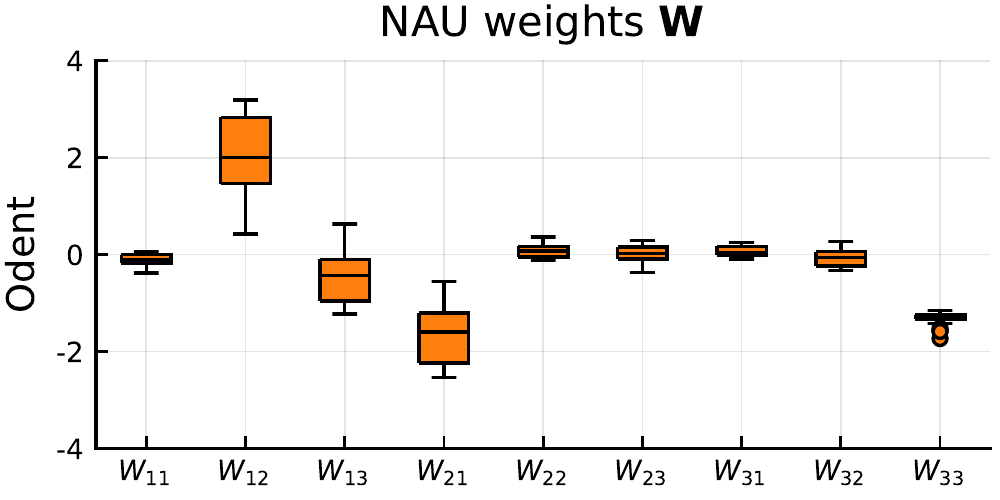}
  \includegraphics[height=.32\linewidth]{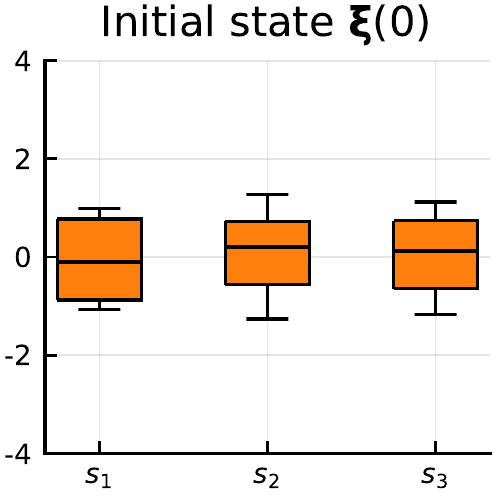}
  \caption{Comparison of the latent distributions for samples from
  Eq.~\eqref{eq:harmonic_samples} of a Rodent (with ARD) and an Odent (without
  ARD). The ODE model is an NAU with three units. Each box represents
  the distribution of one parameter of the ODE. The Rodent, which applies ARD
  to the latent variable clearly pushes all irrelevant parameters to zero.}
  \label{fig:rodent_dists}
\end{figure}

An ideal identification of the harmonic oscillator would be the discovery of
Eq.~\eqref{eq:harmonic_system}. Assuming we know that the system can be described
by a linear ODE, we can represent it by a 2$\times$2-NAU layer.  Thus, we
want to learn an encoder that outputs both the NAU parameters $\bm W$, the
initial position $\xi(0)$ and the velocity $\dot\xi(0)$ for a given trajectory
of observed positions
\begin{equation}
  \x = [\xi_1,\ldots,\xi_K].
\end{equation}
This means that we have six parameters of which
only four are relevant: "1", "$\omega^2$", and the two initial conditions
$\xi(0)$ and $\dot{\xi}(0)$.
To demonstrate the strength of ARD, we over-parameterize the problem with a
third-order ODE (equivalent to a 3$\times$3-NAU):
\begin{align}
  \label{eq:harmonic_ode_model}
  f(\bm\theta,t) \approx \text{NAU}(\bm\xi,t) = \bm W \bm\xi(t)
\end{align}
This third order ODE has twelve parameters in total, of
which still only four are relevant.  The observation operator $H$ is defined
such that only the first component of the state enters into the ELBO:
\begin{equation}
   \psi(\bm z) = H \cdot \psi(\bm\theta, \bm\xi(0), \bm t)
               = [1,0,0] \cdot \psi(\bm\theta, \bm\xi(0), \bm t).
\end{equation}
To demonstrate the advantage of the ARD prior, we train a Rodent and its
variant lacking the ARD prior to reconstruct harmonic signals sampled from
\begin{align}
  \label{eq:harmonic_samples}
  p(\x|\omega, \alpha_0, \se) = \N(\x | \sin(\omega \bm{t} + \alpha_0), \se),
\end{align}
where $p(\alpha_0) = \U(0, 2\pi)$ and $p(\omega) = \U(0.5, 3)$. For both Rodent and
Odent the dense part of the encoder (which predicts $\bm \xi(0)$) has two relu
layers with 50 neurons. The convolutional part (responsible for $\bm W$) has
three convolutional layers with 16 channels and a filter size of 3$\times$1.

\paragraph{The identified manifolds}%
\label{par:the_identified_manifolds}
of ODEs that are represented by the learnt latent distributions of Rodent and
Odent are shown in Fig.~\ref{fig:rodent_dists}. Both Rodent and Odent learn a
reduced structure of the latent space, but we can clearly see that the pruning
is much more effective in the Rodent.  It keeps only four relevant parameters
with the rest being almost exactly zero, while without ARD, seven parameters
remain and the rest is not as close to zero.  ARD penalizes large variances in
its prior $\N(0,\laz)$ which provides the increased pressure towards zero.  The
Odent has some, but less, pressure towards zero because of the fixed standard
normal prior $\N(0,1)$.

\begin{figure}
  \centering
  \includegraphics[width=\linewidth]{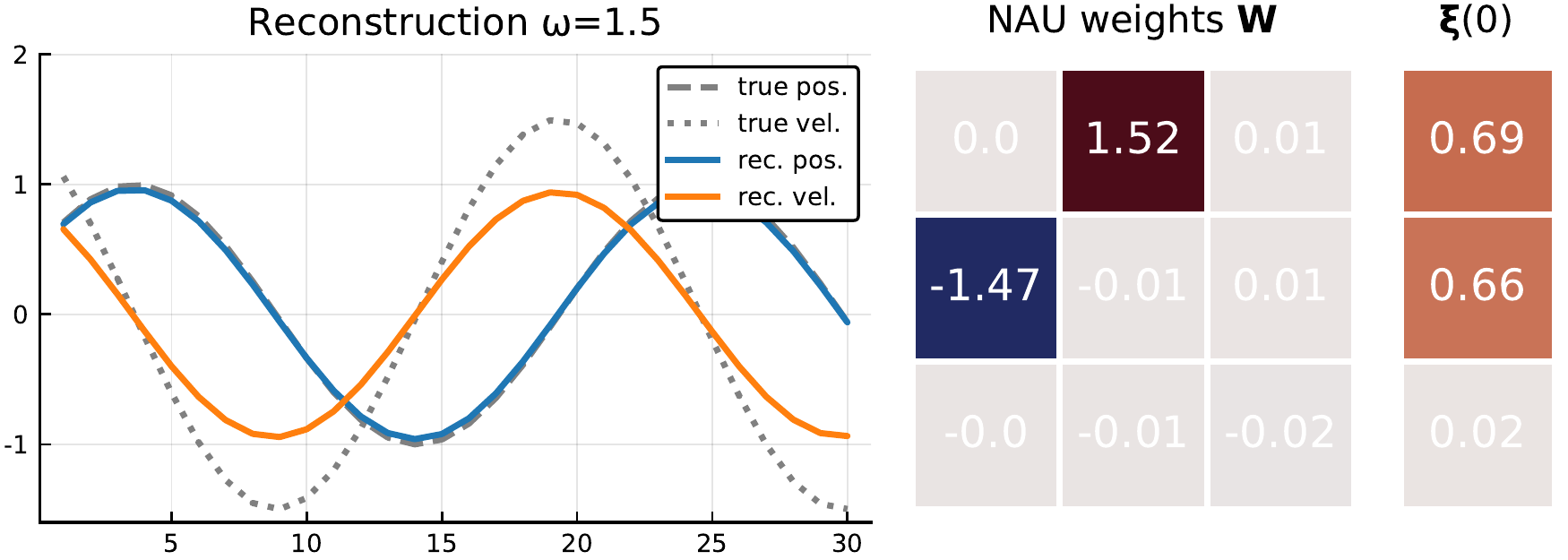}
  \caption{The left plot shows an exemplary input sequence $\x$ (dashed) and
  its reconstruction (blue). The reconstruction is obtained by solving the ODE
  defined by the parameters shown the two plots on the right.  Our framework
  identified a sparse manifold that is very close to the desired solution shown
  in Eq.~\eqref{eq:harmonic_system}. The second state variable differs from the
  true velocity by a factor $b=1/1.52$, which comes from an additional degree of
  freedom that is introduced by the partial observation.}
  \label{fig:harmonic_nau}
\end{figure}

An exemplary Rodent reconstruction and the latent mean corresponding to a
harmonic signal with frequency $\omega=1.5$ is shown in
Fig.~\ref{fig:harmonic_nau}. The latent codes resemble the desired result
shown in Fig.~\ref{fig:perfect_harmonic_nau}.  The position trajectory matches
up nicely, but the second state variable does not exactly represent the
velocity.  Substituting variables for the four relevant parameters in
Eq.~\eqref{eq:transform_W} and Eq.~\eqref{eq:transform_xi} we can write $\bm W^{*}
= [0 \; w_1; w_2 \; 0]$, and $\bm\xi^*=[s_1,s_2]$ (where we omitted the third,
irrelevant row). It follows that
\begin{align}
  w_1 &= 1/b & s_1 &= \xi \\
  w_2 &= -\omega^2 b  & s_2 &= b\dot\xi
\end{align}
where in the case of Fig.~\ref{fig:harmonic_nau}: $b=0.65$ and $\omega=1.49$,
which is close to the original frequency of $\omega=1.5$.
Apart from the additional degree of freedom in $b$, which comes from the
partial observation of the ODE state, the Rodent successfully identified the
simplest solution to the harmonic oscillator.

\begin{figure}
  \centering
  \includegraphics[width=\linewidth]{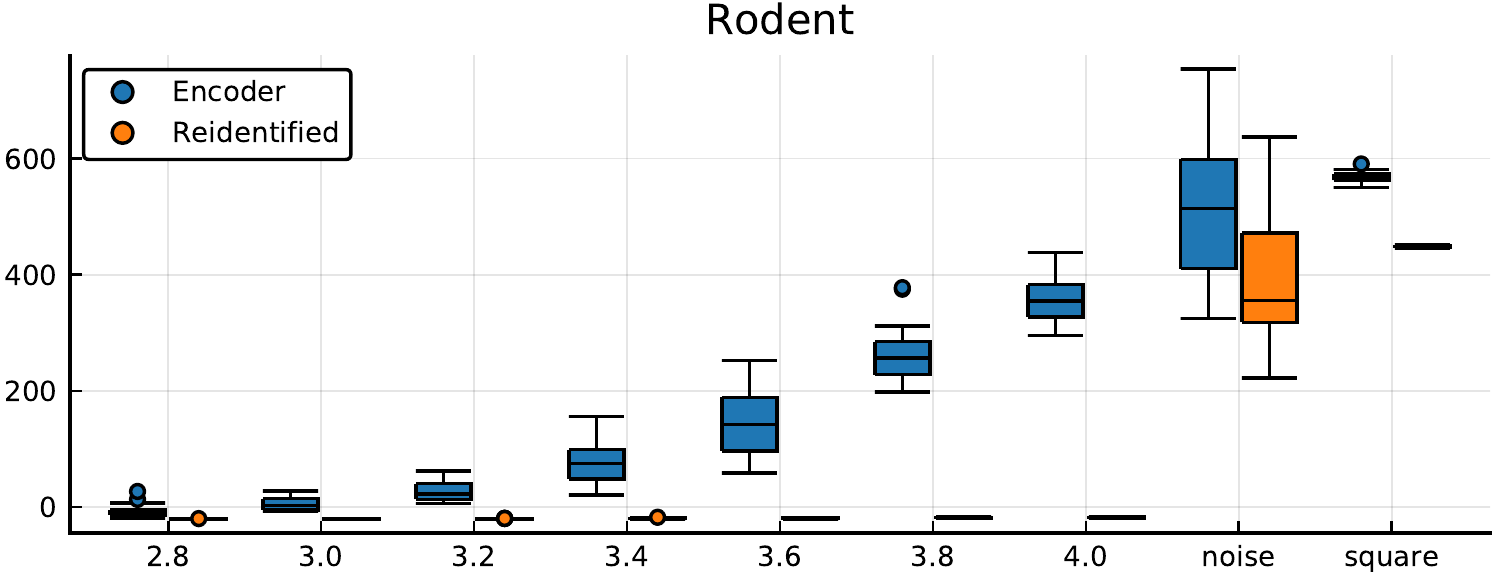}
  \caption{Reconstruction error of time-series that are generated by different
  models: sine waves of frequency $\omega=\{2.8,\ldots, 4.0$\}, (denoted by the
  frequency value on x-axis), square signal and Gaussian noise.
  Reconstruction error based on parameters from the encoder in blue,
  error based on reidentified parameters in yellow.}
  \label{fig:rodent_scores}
\end{figure}

\paragraph{Extrapolation}%
\label{par:extrapolation}
To demonstrate that the Rodent has learnt the manifold of harmonic signals,
Fig.~\ref{fig:rodent_scores} shows the reconstruction error of harmonic samples
with frequencies $[2.8, 3.0, \ldots, 3.8,4]$, which except the frequency
$2.8$ are outside frequencies used to generate training data, random Gaussian
noise, and a square signal of a frequency within the trained
range.  Blue boxes represent the error of samples with the
latent variables as proposed by the encoder. Notice that their error quickly
increases as the frequency increases beyond the trained frequency range (i.e.
over 3).  If we \emph{reidentify} the parameters, the error drops to very low
values, which means that the Rodent can extrapolate far beyond its training
range.  Higher errors for Gaussian noise and square signals stay much higher
than those of harmonic signals which indicates that these signals are outside
the manifold. This means that the model of harmonic signals identified by the
Rodent is correct.

\begin{figure}
  \centering
  \includegraphics[width=\linewidth]{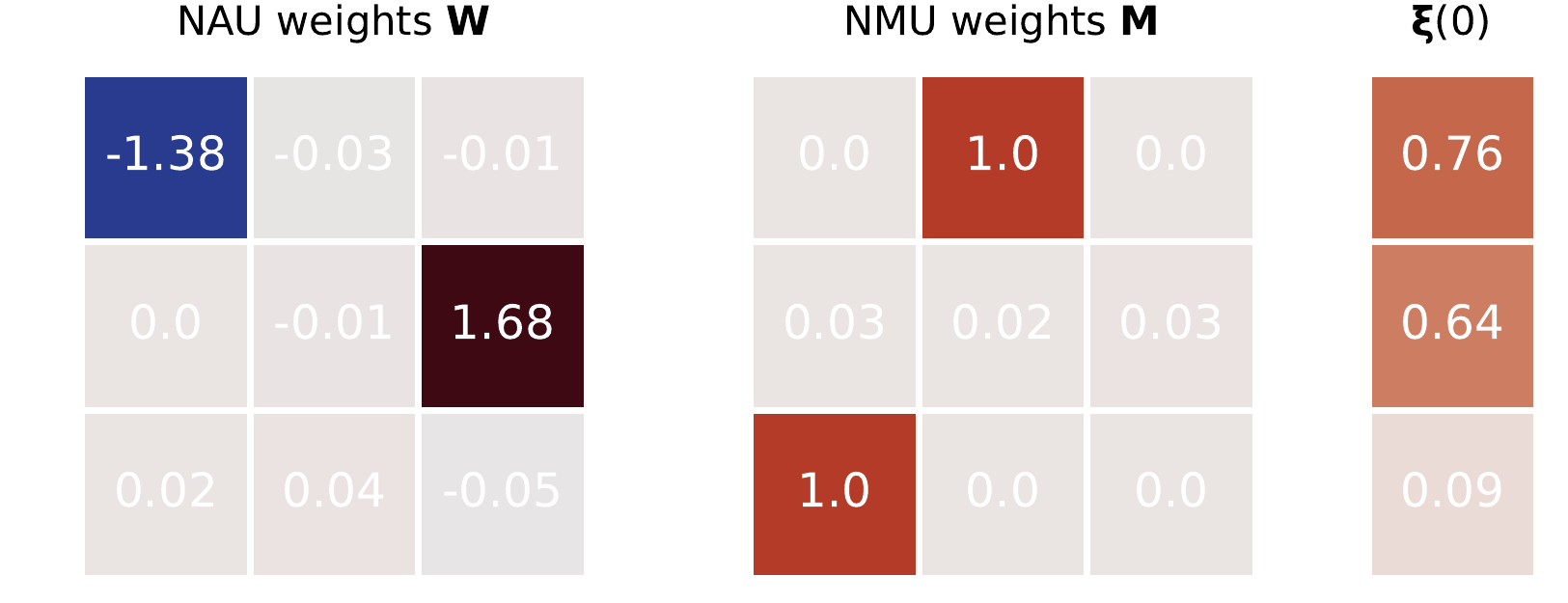}
  \caption{Latent space of the Rodent with chained NMU and NAU
  (Eq.~\eqref{eq:nmu_nau_ode}) that is trained to reconstruct harmonic signals.
  In the example the generating frequency is $\omega=2.5$.}
  \label{fig:harmonic_nmu_nau_mean}
\end{figure}

\paragraph{Adding the NMU}%
\label{par:nmu}
extends the capabilities of the Rodent from learning linear ODEs to nonlinear
ODEs with products of state variables. For the harmonic oscillator this is of
course not necessary, but just to demonstrate that ARD is powerful enough to
prune the weights of the excess NMU we show it in
Fig.~\ref{fig:harmonic_nmu_nau_mean}.  The ODE model then reads:
\begin{align}
  \label{eq:nmu_nau_ode}
  f(\bm\theta,t) \approx \text{NAU}_{\bm\theta}(\text{NMU}_{\bm\theta}(\bm\xi(t)))
\end{align}
The NMU just picks out two state variables such that the hidden activation
becomes $\bm h = [s_2,1,s_1]'$, which is then passed to the NAU with the same
result as before.

\subsection{Superimposed harmonic oscillators}%
\label{sub:superimposed_harmonic_oscillators}

\begin{figure}
  \centering
  \includegraphics[width=\linewidth]{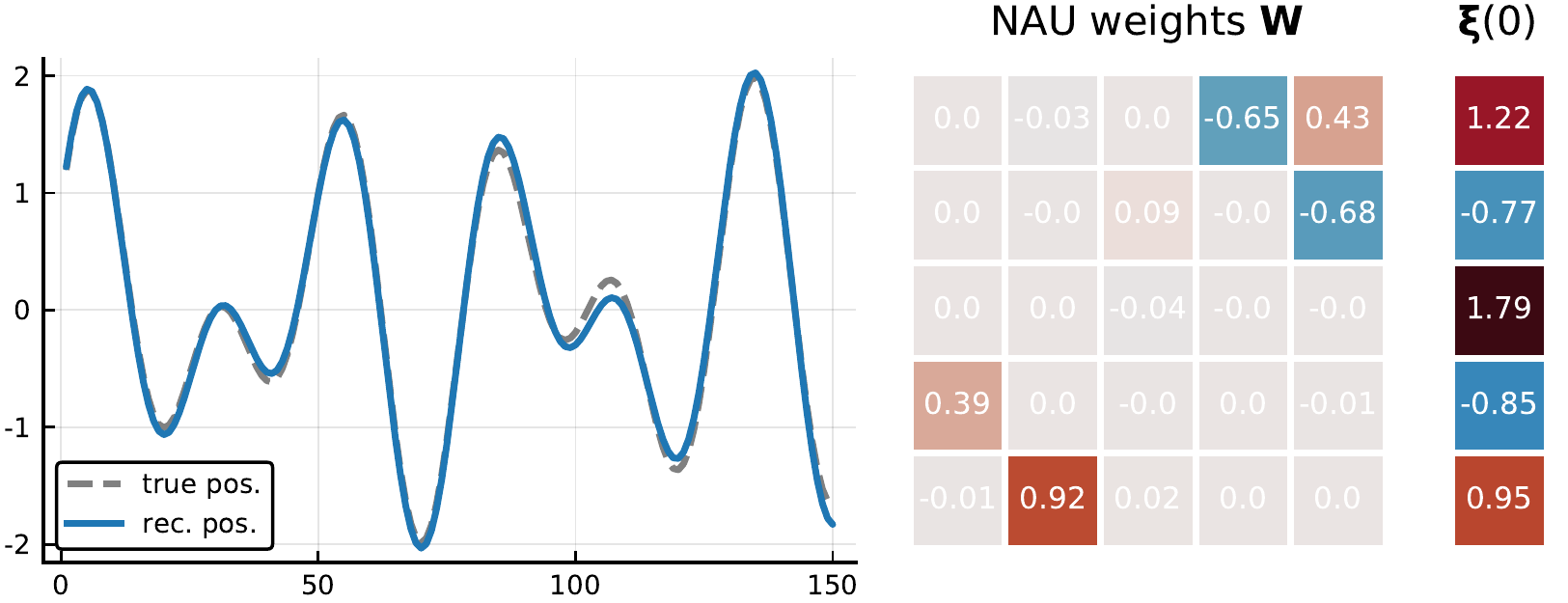}
  \caption{Rodent learning superimposed oscillators with observation operator
  $H=[1,0,0,0,0]$, which is too minimal to force the Rodent to learn two
  harmonic oscillators as basis functions. The summation of two states in the
  observed first state variable is just by chance.}
  \label{fig:double_harmonic_nau_h10000}
\end{figure}

\begin{figure}
  \centering
  \includegraphics[width=\linewidth]{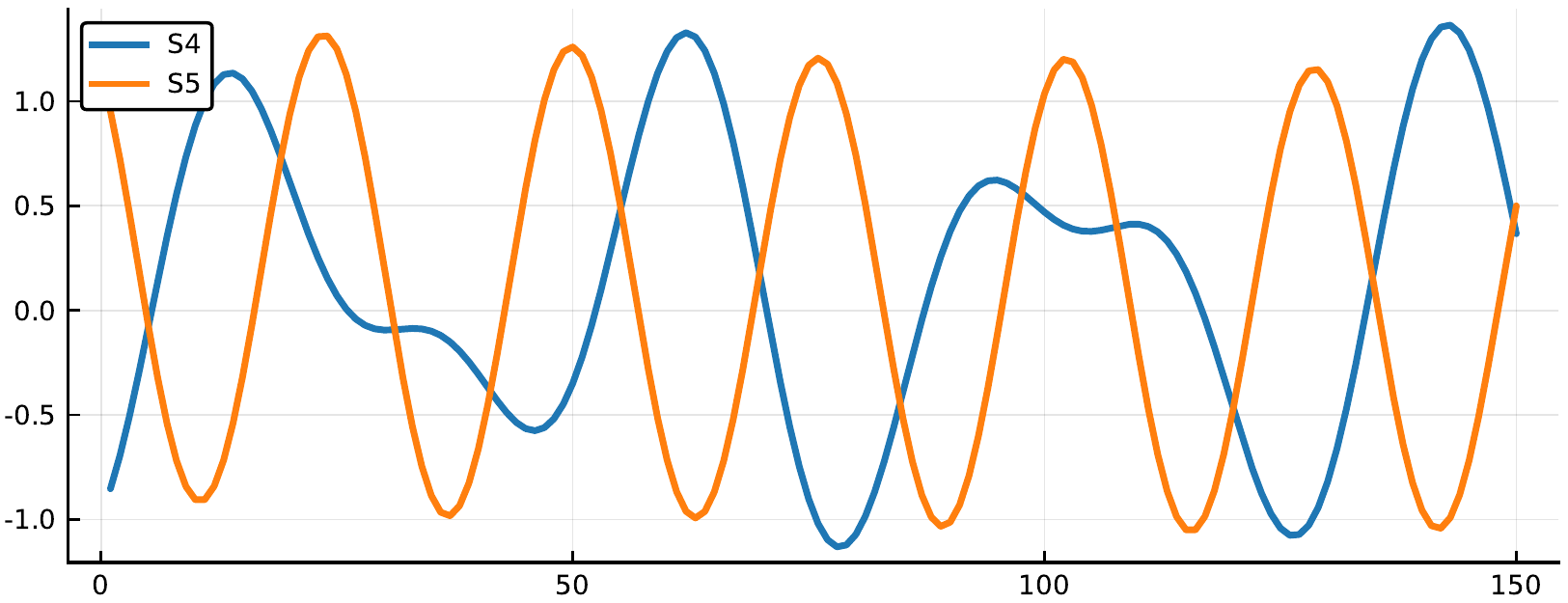}
  \caption{State trajectories of the two states that are summed by the Rodent
  with a 5$\times$5-NAU. One of them approximately harmonic, the other more
  complicated. The Rodent did not decouple the superimposed signal.}
  \label{fig:double_harmonic_nau_h10000_states}
\end{figure}

\begin{figure}
  \centering
  \includegraphics[width=\linewidth]{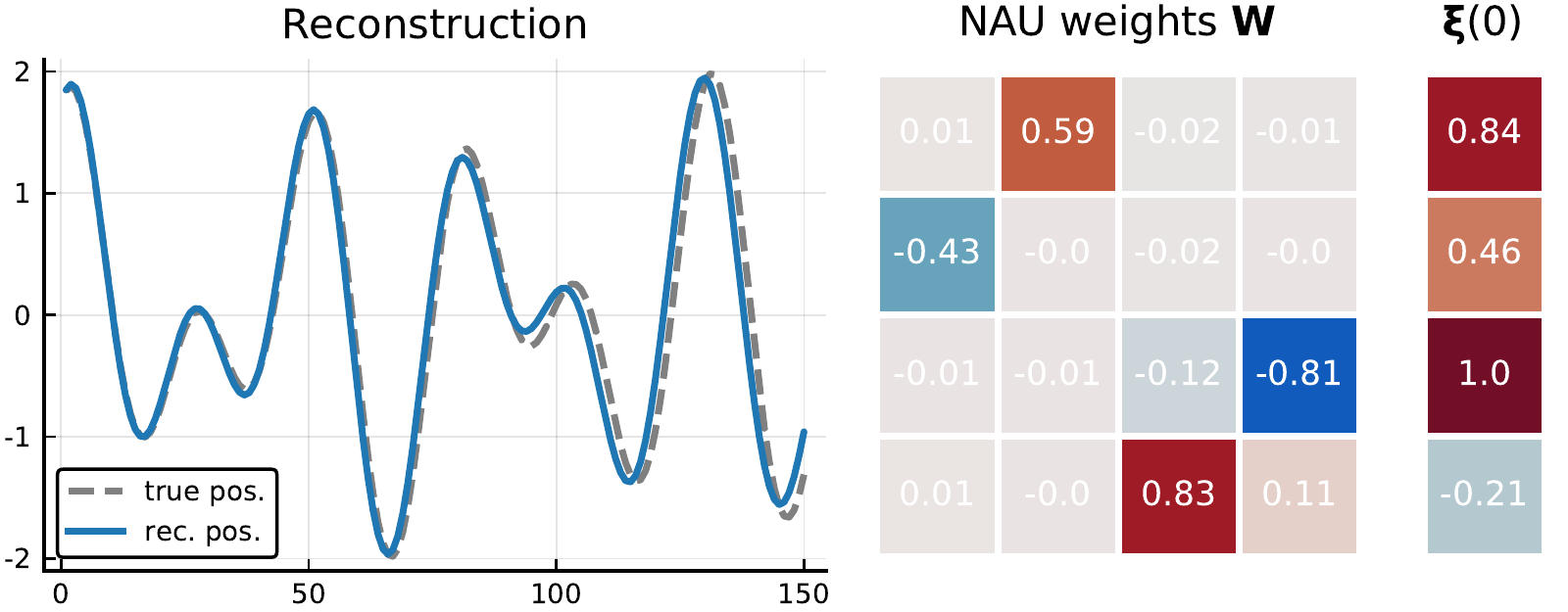}
  \caption{Rodent with 4$\times$4-NAU and $H=[1,0,0,0]$, which learns to
  decouple the superimposed inputs. The two decoupled oscillators are visible
  in the latent means on the right.}
  \label{fig:double_harmonic_nau}
\end{figure}

\begin{figure}
  \centering
  \includegraphics[width=\linewidth]{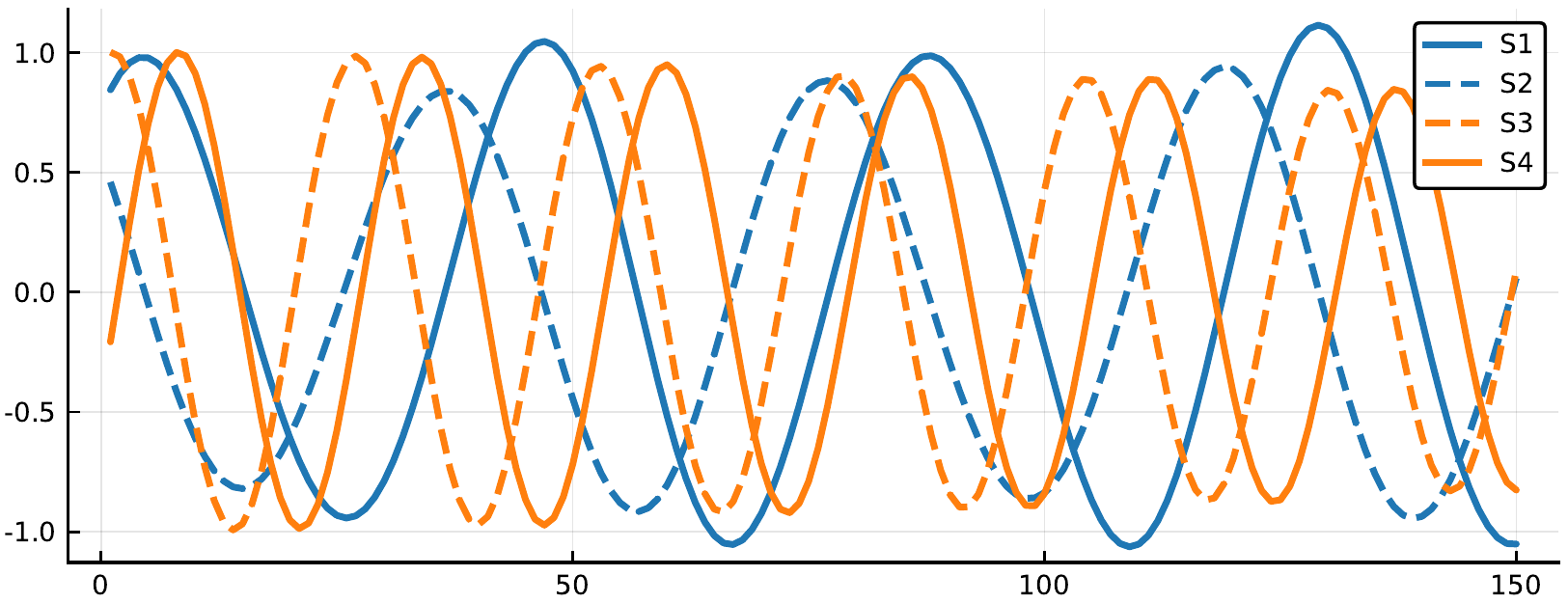}
  \caption{State trajectories of the Rodent with a 4$\times$4-NAU.  The two
  yellow lines represent position and transformed derivative of the first
  oscialltor, the blue lines represent the second oscillator respectively.}
  \label{fig:double_harmonic_nau_states}
\end{figure}

The influence of different observation operators can be nicely demonstrated by
letting the Rodent identify two superimposed harmonic oscillators.  We sample
time-series from
\begin{equation}
  \label{eq:double_harmonic_dist}
  \nonumber
  p(\bm x|\alpha_0, \sigma_e)
    = \mathcal{N}(\bm x| \sin(\omega_1(\bm t + \Delta t))
                       + \sin(\omega_2(\bm t + \Delta t)), \sigma_e),
\end{equation}
where $\omega_1 = 0.5$, $\omega_2=0.8$, and $p(\Delta
t)=\mathcal{U}(0,2\pi/\omega_1)$.  Assume we want to learn a system with two
oscillators with minimal observation operator that only sees the superimposed
position ($H = [1, 0, 0, 0, 0]$).  In this case we will need an NAU with at
least four states (two for each oscillator) and another state which contains
their sum. Hence we train a Rodent with a 5$\times$5-NAU, which results in a
latent structure as shown in Fig.~\ref{fig:double_harmonic_nau_h10000}.
A crude calculation based on the previous experiment would
suggest two parameters in $\bm W$ for each oscillator plus two for their summation.
Surprisingly the number of relevant parameters is five, so ARD found a way of
removing another parameter.
Further, the Rodent does learn a sum of two state variables $s_4$ and $s_5$ (row one of $\bm
W$), but the rest of the NAU does not look like two oscillators. In fact, as we
can see in Fig.~\ref{fig:double_harmonic_nau_h10000_states} it seems like the
Rodent learnt a superposition of one harmonic oscillator and another, more
complicated trajectory. This demonstrates that the \emph{simplest} solution,
in terms of the number of parameters, is not always the most \emph{interpretable}
one. We gain no advantage from learning a superposition of a sinus and something
that is essentially as complicated as the input data, although ARD reduced the
number of parameters even further than expected.
With a slightly more restrictive observation operator that sums two arbitrary
state variables (here $H = [1, 0, 1, 0]$ we can encourage the Rodent to learn a
superposition of harmonic oscillators. As $H$ takes over the summation, we can
also downsize the NAU to 4$\times$4.  Fig.~\ref{fig:double_harmonic_nau} shows
the latent mean, which resembles two oscillators\footnote{The
oscillators are of course possibly transformed just as in
Sec.~\ref{sub:identification_of_the_harmonic_oscillator}}.
Plotting all state trajectories (Fig.~\ref{fig:double_harmonic_nau_states})
reveals that the Rodent now indeed learns to decouple the superimposed input
signal.  The two yellow lines show position (solid line) and its transformed
derivative (dashed line) of the first oscillator, and the two blue lines show
the same for the second oscillator.
This experiment concludes that the simplicity that ARD enforces does
not necessarily align with our human idea of simplicity, which favours the
decomposable result.

\subsection{Lotka-Volterra System}%
\label{sub:lotka_volterra_system}

A system that requires the nonlinear capabilities of the Rodent is the
Lotka-Volterra (LK) system:
\begin{align}
  \label{eq:lotka_volterra_system}
  \dot x &= \alpha x - \beta xy\\
  \dot y &= -\delta y + \gamma xy
\end{align}
It describes the dynamical evolution of a predator population $y$ and a prey
population $x$.  To obtain one input sample $\x$ we solve the LK system for
$N=600$ timesteps $\Delta t=0.1$ with parameters that are sampled from
\begin{align}
  \label{eq:lotka_dists_definition}
   p(\alpha) &= \mathcal{U}(2,2.5)
  &\beta   = 1 \\
   p(\delta) &= \mathcal{U}(3,3.5)
  &\gamma  = 1
\end{align}
and pick out a random starting point from within the resulting sequence.
In this example, we observe the full state with the operator $H$:
\begin{equation}
  \psi(\bm z) = \text{vec}([1,1] \cdot \psi(\bm\theta, \bm\xi(0), \bm t)).
\end{equation}
\begin{figure}
  \centering
  \includegraphics[width=\linewidth]{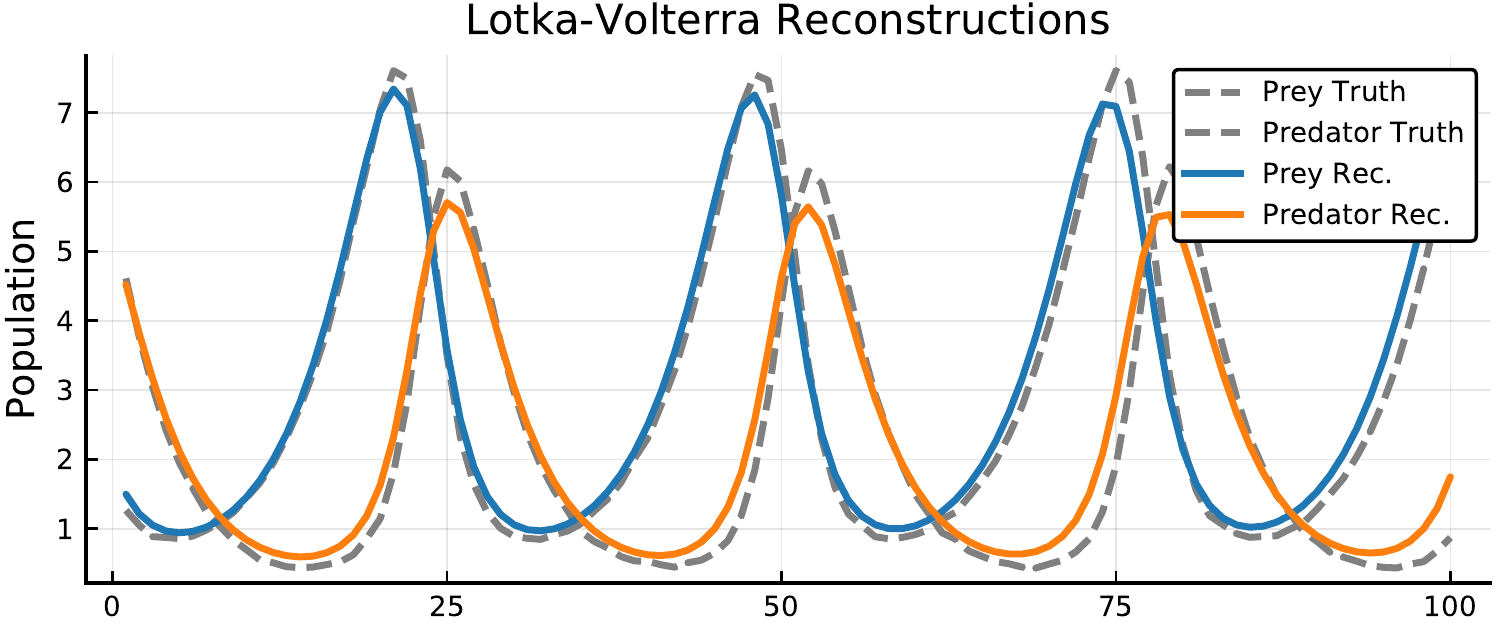}
  \caption{True evolution of the LK system compared to reconstructions created
  by the Rodent. Corresponding latent means in Fig.~\ref{fig:lotka_nmu_nau_mean}.}
  \label{fig:lotka_nmu_nau_recs}
\end{figure}
\begin{figure}
  \centering
  \includegraphics[width=\linewidth]{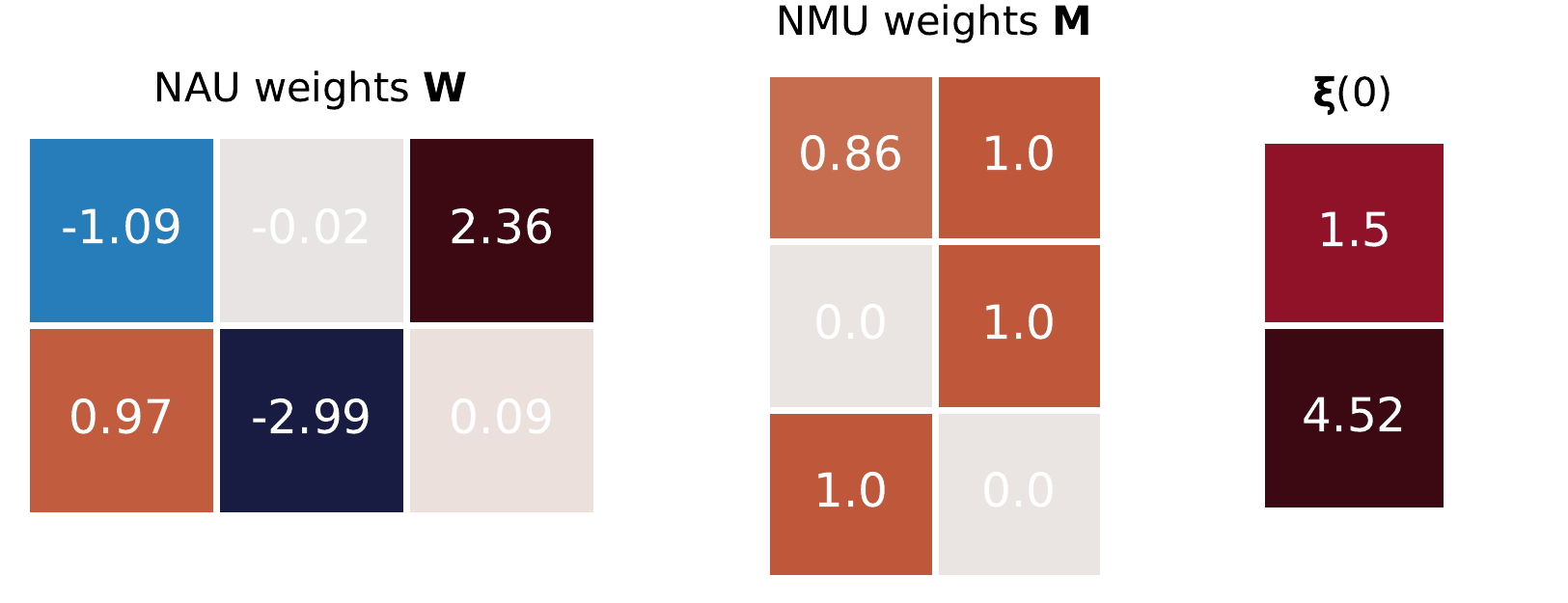}
  \caption{Latent means of an NAU/NMU chain that decodes into an LK sequence.
  The NMU weights can be read as a matrix multiplication where all sums are
  replaced by products. It performs the necessary multiplications, while the
  NAU weights represent the LK parameters $\alpha,\beta,\delta$, and $\gamma$.
  Corresponding reconstruction in Fig.~\ref{fig:lotka_nmu_nau_recs}.}
  \label{fig:lotka_nmu_nau_mean}
\end{figure}

Fig.~\ref{fig:lotka_nmu_nau_recs} shows the reconstructions of the Rodent
for both predator and prey populations.
Fig.~\ref{fig:lotka_nmu_nau_mean} depicts the ODE that produced
the reconstruction.
The NMU in the first layer of the ODE creates the necessary products
($xy,x,y$) which are then summed by the second NAU layer to produce the correct
ODE. The resulting parameters are very close to the optimal solution in
Eq.~\eqref{eq:lotka_volterra_system}.

\section{Conclusion}%
\label{sec:conclusion}

We have shown how to discover the governing equations of a subset of nonlinear
ODEs purely from partial observations of the ODE's state. This was done by extending
the VAE with ARD and by replacing the decoder by an ODE solver, which results
in the relevant ODE identifier (Rodent).
Our approach results in a highly interpretable latent space containing physical
meaning from which we can read out the governing equation of the analysed
system.
We demonstrated that it is possible to learn not only a single model for a given
system, but a manifold that represents a more general class of systems.  After
the correct manifold is identified, it is possible to extrapolate far beyond the
training range, which is another hint at the Rodent's ability to extract
physical concepts from data.
Extending the VAE by ARD lead to a significant increase in the sparsity of the
latent space, which is key to an interpretable result.
Our approach reaches its limits as soon as the observation operator is too
loose. Such an operator introduces degrees of freedom in the system that impair
the Rodent's ability to distil data into a form that is \emph{simple} for
humans.

A promising extension of the Rodent would be to split the ODE model into a
shared and a specific part. The example of the harmonic oscillator contains
variable parameters, namely $\omega$ and the initial state $\bm\xi(0)$, as well
as a constant parameter ``1''. It would be possible to learn an ODE that is
fixed for all samples and let the encoder output an ODE that is specific for
each sample
\begin{align}
  \label{eq:const_spec}
  f(\bm\theta,t) \approx g(\bm\theta_{\text{shared}}) + h(\bm\theta_{\text{spec}}).
\end{align}

Another obvious extension of our model would be to increase the richness of the
ODEs that can be learnt in the latent space of the Rodent.  This would be
possible by using the extended NALU from Sec.~\ref{sec:neural_arithmetic} that
can learn arbitrary power functions. To do this, we would need to solve
the problem of recovering the sign of the input in the multiplication layer,
which might also make convergence more stable.  A further extension to partial
differential equations will be explored in future work.

\section{Acknowledgments}
 Research presented in this work has been supported by the Grant Agency of the
 Czech Republic no. 18-21409S. The authors also acknowledge the support of the
 OP VVV MEYS funded project CZ.02.1.01/0.0/0.0/16\_019/0000765 ``Research Center
 for Informatics''.

\bibliography{main}
\bibliographystyle{icml2020}

\end{document}